\documentclass[conference]{IEEEtran}
\IEEEoverridecommandlockouts
\usepackage{cite}
\usepackage{amsmath,amssymb,amsfonts}
\usepackage{algorithmic}
\usepackage{graphicx}
\usepackage{textcomp}
\usepackage[table]{xcolor}

\def\BibTeX{{\rm B\kern-.05em{\sc i\kern-.025em b}\kern-.08em
    T\kern-.1667em\lower.7ex\hbox{E}\kern-.125emX}}

\usepackage{booktabs}
\usepackage{tabularx}   
\usepackage{multirow}
\usepackage{makecell}

\usepackage{subcaption} 

\newtheorem{assumption}{Assumption}
\newtheorem{theorem}{Theorem}
\newtheorem{lemma}{Lemma}

\begin{document}

\title{CA-HFP: Curvature-Aware Heterogeneous Federated Pruning with Model Reconstruction}

\author{Gang Hu, Yinglei Teng, Pengfei Wu, and Shijun Ma}

\maketitle

\begin{abstract}
Federated learning on heterogeneous edge devices requires personalized compression while preserving aggregation compatibility and stable convergence. We present Curvature-Aware Heterogeneous Federated Pruning (CA-HFP), a practical framework that enables each client perform structured, device-specific pruning guided by a curvature-informed significance score, and subsequently maps its compact submodel back into a common global parameter space via a lightweight reconstruction. We derive a convergence bound for federated optimization with multiple local SGD steps that explicitly accounts for local computation, data heterogeneity, and pruning-induced perturbations; from which a principled loss-based pruning criterion is derived. Extensive experiments on FMNIST, CIFAR-10, and CIFAR-100 using VGG and ResNet architectures under varying degrees of data heterogeneity demonstrate that CA-HFP preserves model accuracy while significantly reducing per-client computation and communication costs, outperforming standard federated training and existing pruning-based baselines.
\end{abstract}


\section{Introduction}
\label{sec:intro}

Federated learning (FL) enables a collection of distributed devices to collaboratively train a shared model under the coordination of a central server while keeping raw data local \cite{Survey_Advances_and_open_problems_in_FL}. This paradigm is particularly attractive in cross-device settings—e.g., mobile phones, IoT sensors, and edge gateways—where data are naturally decentralized and privacy or regulatory constraints prohibit central aggregation. In practice, however, deploying FL at the edge faces severe resource constraints: many devices have limited compute, memory, battery, and network bandwidth, and wireless channels are often unreliable\cite{Survey_FL_in_MEN_yangJ,IEEEWC_FL_over_WC}. These limitations make it difficult for all clients to perform full local training or to exchange dense model updates efficiently.

Another fundamental challenge in FL stems from heterogeneity across participating clients\cite{Survey_FL_in_MEN_yangJ}, which poses severe obstacles to practical deployment. First, the disparities in clients’ computation and communication capabilities, i.e., \emph{system heterogeneity}\cite{FedProx}, results in straggling devices and unstable participation, making it impractical to require all clients to perform full-model training. Second, \emph{non-independent and non-identically distributed} (non-IID) data across clients lead to divergent local updates and degrade global convergence and generalization. Factually, these two types of heterogeneity are tightly coupled in resource-constrained environments, i.e., devices with limited resources often own small, biased datasets, and the resulting non-uniform local training further amplifies optimization instability. Therefore, designing FL algorithms that are simultaneously communication- and computation-efficient, and robust to heterogeneity remains a fundamental problem.

To alleviate the high communication and computation costs induced by resource constraints and heterogeneity in FL, extensive efforts have investigated model pruning–based FL frameworks, which reduce model size by training and transmitting sparse or compact submodels. Representative approaches include adaptive and distributed pruning mechanisms such as PruneFL \cite{PruneFL} and FedMP\cite{FedMP}, which tailor pruning strategies to heterogeneous device capabilities, as well as more recent frameworks that incorporate server–client collaboration or multi-stage pruning to further improve efficiency, e.g., Complement Sparsification\cite{Complement_Sparsification}, Dual Model Pruning\cite{Dual_Model_Pruning}, and FedPE \cite{FedPE}. Other works extend pruning to address personalization, domain shift, privacy, or multi-task learning, such as DapperFL, Fed-LTP, and FedLPS\cite{DapperFL,Fed_LTP,FedLPS}. Despite notable efficiency gains, current FL-with-pruning approaches still struggle to jointly achieve communication efficiency, aggregation robustness, and stable convergence under coupled system and data heterogeneity.

\begin{figure*}[t]
\centering
\includegraphics[width=15cm,height=5.5cm]{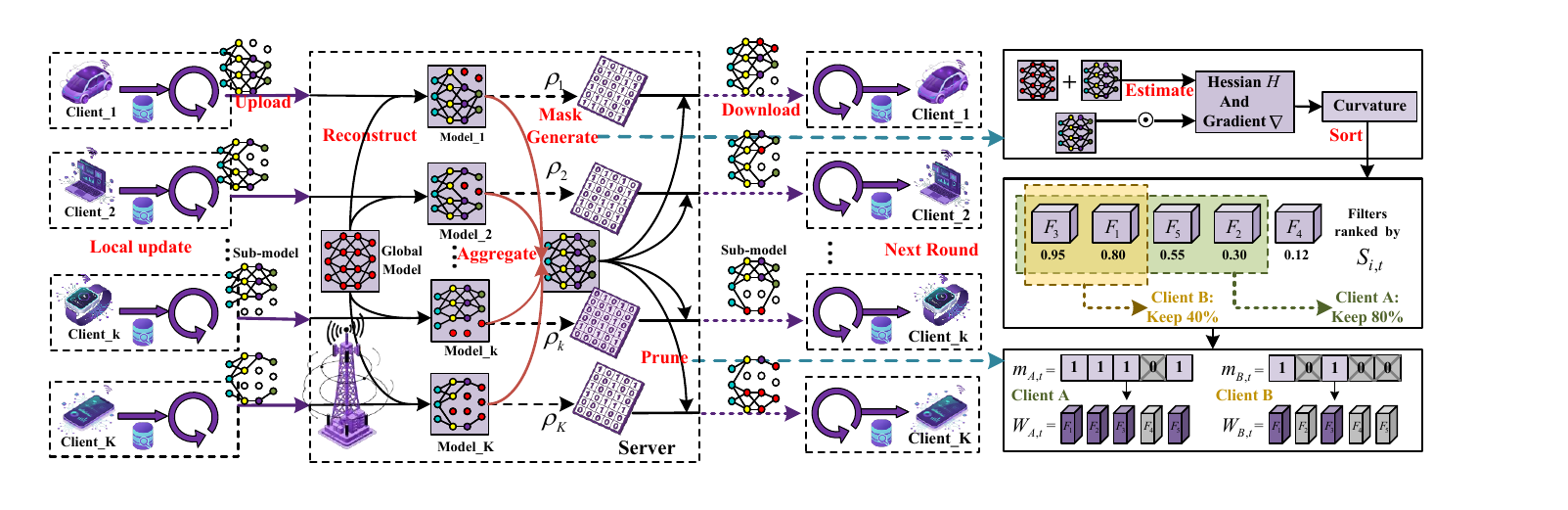}
\vspace{-3mm}
\caption{Curvature-Aware Heterogeneous Federated Pruning (CA-HFP) Framework.}
\vspace{-5mm}
\label{HFLP}
\end{figure*}

In this paper, we investigate heterogeneous federated learning with personalized, structured pruning for resource-constrained edge environments. Our objective is to let each client adaptively compress its local model to satisfy device constraints while still contributing meaningful, aggregation-compatible updates to the global model and preserving convergence guarantees. To this end, we introduce \emph{Curvature-Aware Heterogeneous Federated Pruning (CA-HFP)}, a practical framework that (i) lets clients adopt personalized structured pruning ratios guided by a principled importance metric, (ii) maps each compact client submodel back into a common global parameterization via a lightweight server-side reconstruction prior to aggregation, and (iii) is supported by a theoretical analysis that quantifies how local computation, client dissimilarity, and pruning perturbations jointly affect convergence. Specifically, the main contributions of this work are summarized as follows:
\begin{itemize}
\item We propose CA-HFP, a heterogeneous FL framework that supports personalized structured pruning to accommodate resource-constrained devices with diverse computation and communication capabilities.

\item We provide a convergence analysis for federated optimization with personalized pruning, explicitly quantifying the effects of local computing, data heterogeneity, and pruning-induced perturbations. Then a pruning criterion and a curvature-aware loss perturbation are derived to guide HFP to prune unimportant parameters, which mitigates pruning-induced aggregation bias and improving robustness under non-IID data distributions.

\item We develop a reconstruction-based aggregation mechanism that resolves structural mismatch among heterogeneous pruned sub-models, enabling unified synchronous aggregation in the original parameter space.

\item We conduct extensive experiments across multiple datasets and model architectures, demonstrating that CA-HFP consistently outperforms baselines in terms of accuracy, efficiency, and robustness to resource-constrained devices. Ablation studies further validate the effectiveness of the proposed reconstruction mechanisms.
\end{itemize}


\section{System Model} \label{Section-HFLP}
In this section, we propose a heterogeneous federated pruning framework that effectively improves both communication and computation efficiency. We then analyze the convergence of the proposed federated learning algorithm with $E$ local SGD steps and model pruning. In particular, we explicitly characterize the impact of local computing, data heterogeneity, and model pruning on the global convergence behavior.

\subsection{Heterogeneous Federated Pruning Framework}
We consider a standard federated optimization problem of learning a $d$-dimensional model $w \in \mathbb{R}^d$ from $K$ distributed clients:
\begin{equation}
    \mathop {\min }\limits_{w \in {\mathbb{R}^d}} F(w): = \sum\nolimits_{k = 1}^K {{p_k}{F_k}(w)} ,
    \label{eq:global_obj}
\end{equation}
where $F_k(w) = \mathbb{E}_{\xi\sim\mathcal{D}_k}[f(w;\xi)]$ is the local expected loss of client $k$, $\mathcal{D}_k$ is its data distribution, and $p_k = n_k / \sum_{j=1}^K n_j$ denotes the relative weight proportional to the local data size $n_k$. These clients iteratively train the local sub-model and aggregate the global full-model through multiple rounds. This framework is illustrated in Fig.~\ref{HFLP}, and each round consists of the following three stages.

\emph{(i) Model pruning and downloading:}
At communication round $t \! = \! 0,1,\dots,T-1$, the server maintains a global full model $w_t$ and interacts with all clients. To account for device and channel heterogeneity, each participating client $k$ is associated with a binary pruning mask $m_{k,t} \in \{0,1\}^d$ that specifies which parameters of the global model are activated on that client. The personalized local sub-model distributed to client $k$ is thus
\begin{equation}
    w_{t}^{(k,0)} = m_{k,t} \odot w_t,
    \label{eq:masked_init}
\end{equation}
where $\odot$ denotes the Hadamard (element-wise) product. 

\emph{(ii) Local model updating and uploading:}
Given the personalized sub-model $w_t^{(k,0)}$, client $k$ performs $E$ local steps of stochastic gradient descent (SGD) on its local dataset:
\begin{equation}
    w_t^{(k,\tau\!+\!1)} 
   \! = w_t^{(k,\tau)}\! - \!\eta \nabla f\big(w_t^{(k,\tau)};\xi_t^{(k,\tau)}\big),
    \, \tau \! = \! 0,\dots,E\!-\!1,
    \label{eq:local_sgd}
\end{equation}
where $\eta > 0$ is the local step size and $\xi_t^{(k,\tau)}$ is a mini-batch sampled from $\mathcal{D}_k$. Note that the gradient update inEq. ~\eqref{eq:local_sgd} is implicitly restricted to the active parameters specified by $m_{k,t}$. After $E$ local steps, client $k$ obtains its updated sub-model $w_t^{(k,E)}$ and uploads it to the server. 

\emph{(iii) Reconstruction and synchronous aggregation:}
Due to the heterogeneous pruning masks and resulting sub-model structures, it is not feasible to directly apply conventional FedAvg aggregation. To enable a unified aggregation in the original parameter space, the server first reconstructs the local models to the full dimension by expanding the missing entries according to the current global model $w_t$. Denoting the reconstructed full-dimensional model of client $k$ by $\tilde{w}_t^{(k,E)}$, the server performs a weighted synchronous aggregation:
\begin{equation}
    {w_{t + 1}} = \sum\nolimits_k {{p_k}\tilde w_t^{(k,E)}} ,
    \label{eq:agg}
\end{equation}

The aggregated iterate ${w}_{t+1}$ can be viewed as an inexact global gradient update. For analysis, it is convenient to rewriteEq. ~\eqref{eq:agg} in the form
\begin{equation}
    {w_{t + 1}} = {w_t} - {e_t} + \sum\nolimits_k {{p_k}\Delta _t^k} ,
    \label{eq:global_update_noise}
\end{equation}
where $\Delta _t^k = w_t^{k,E} - w_t^{k,0}$ and $e_t$ is an noise term capturing the global model loss introduced by personal pruning.

\subsection{Convergence Analysis}



We make the following standard assumptions.

\begin{assumption} (\emph{L-smooth}). \label{as:smooth}
Each local objective $F_k$ is $L$-smooth, i.e., for all $w,u\in\mathbb{R}^d$,
\begin{equation}
    \|\nabla F_k(w) - \nabla F_k(u)\| \le L \|w-u\|.
\end{equation}
Consequently, the global objective $F$ is also $L$-smooth.
\end{assumption}

\begin{assumption} \emph{(Unbiasedness and Bounded variance of stochastic gradients)}. \label{as:var}
For all clients $k$, all $w\in\mathbb{R}^d$, and all $\xi\sim\mathcal{D}_k$,
\begin{equation}
    \mathbb{E}[\nabla \! f_k(w;\!\xi )] \! = \! \nabla \! F_k(\!w\!), \; \mathbb{E} \! \left[ \! {{{\left\| {\nabla \! {f_k}(w; \! \xi ) \!-\! \nabla \! {F_k}(\!w\!)} \right\|}^2}} \! \right] \! \leqslant \! \sigma _k^2.
\end{equation}
\end{assumption}

\begin{assumption} (Bounded client dissimilarity). \label{as:heterogeneity}
There exists $\zeta > 0$ such that, for all $w$,
\begin{equation}
   {\sum\nolimits_k {{p_k}\left\| {\nabla {F_k}(w) - \nabla F(w)} \right\|} ^2} \leqslant {\zeta ^2}.
\end{equation}
\end{assumption}

\begin{lemma}[Result of one round]
    \label{lem:sigma_eff}
    Suppose that each client performs $E$ local SGD steps as in Eq. \eqref{eq:local_sgd}, and the server aggregates according to Eq. \eqref{eq:agg}. Let the local step size $0 < \eta  \leqslant \frac{1}{{4LE}}$, and define $e_t$ implicitly by Eq. \eqref{eq:global_update_noise}. Under Assumptions~\ref{as:smooth}--\ref{as:heterogeneity}, the following bound holds:
    \begin{equation}
        \begin{gathered}
  \mathbb{E}\left[ {F\left( {{w_{t + 1}}} \right)} \right] \leqslant F\left( {{w_t}} \right) - \frac{{\eta E }}{4}\mathbb{E}\left[ {{{\left\| {\nabla F\left( {{w_{t + 1}}} \right)} \right\|}^2}} \right] \hfill \\
   + 4L{\eta ^2}{E^2}\left( {{\sigma ^2} + {\zeta ^2}} \right) + 3L\mathbb{E}\left[ {{{\left\| {{e_t}} \right\|}^2}} \right]. \hfill \\ 
\end{gathered} 
        \label{eq:sigma_eff}
    \end{equation}
\end{lemma}

\begin{theorem}[FL Convergence with personal pruning]
    \label{thm1}
    For any number of communication rounds $T\ge 1$, the iterates $\{w_t\}_{t=0}^{T}$ generated by the federated algorithm with $E$ local SGD steps per round and model pruning satisfy:
    \begin{equation}
        \begin{gathered}
  \mathop {\lim }\limits_{T \to \infty } \frac{1}{T}\sum\limits_{t = 0}^{T - 1} {\mathbb{E}\left[ {{{\left\| {\nabla F\left( {{w_{t + 1}}} \right)} \right\|}^2}} \right]}  = \frac{{4\left( {F\left( {{w_0}} \right) - {F^*}} \right)}}{{\eta ET}} \hfill \\
   + 16L\eta E\left( {{\sigma ^2} + {\zeta ^2}} \right) + \frac{{12L}}{{\eta E}}\frac{1}{T}\sum\limits_{t = 0}^{T - 1} {\mathbb{E}\left[ {{{\left\| {{e_t}} \right\|}^2}} \right]} . \hfill \\ 
\end{gathered} 
\label{eq:theorem1_generic}
    \end{equation}
\end{theorem}

The proof is omitted in this paper due to the limited space. The full proofs can refer to Appendix and online version.

\textbf{Remark 1.} Theorem~\ref{thm1} shows that the proposed CA-HFP algorithm converges to a neighborhood of a stationary point. The first term indicates that increasing either the number of communication rounds $T$ or the number of local steps $E$ accelerates convergence. Moreover, a moderate $E$ can help absorb pruning-induced bias, whereas an excessively large $E$ may amplify the steady-state error caused by gradient noise $\sigma^2$ and client heterogeneity $\zeta^2$. \emph{The result highlights a trade-off among local computation, data heterogeneity, and pruning strength, and suggests that moderate pruning combined with a properly chosen $E$ is important to maintain stable convergence.}

\section{Curvature-Aware Personal Pruning Design}
In this section, we develop a personalized pruning method to minimize the loss degradation induced by model pruning. Furthermore, we present in detail the model reconstruction mechanism designed to address the model heterogeneity caused by personalized pruning, thereby enabling local model aggregation in a FedAvg-like manner.

\subsection{Loss-Perturbation Analysis Induced by Model Pruning}
According to Theorem 1, each pruning operation in round $t$ introduces an additional noise term $e_t$, which degrades the convergence rate and increases the training error. To mitigate this, we aim to minimize the loss error with personalized pruning at each communication round, i.e., we formulate the \emph{loss perturbation} as follows:
\begin{equation}\label{loss_error}
  F\left( {{w_t} - {e_t}} \right) - F\left( {{w_t}} \right).
\end{equation}
According to the second-order Taylor expansion, the FL loss with model pruning can be reformulated as follows:
\begin{equation}\label{Taylor_loss_error}
  F\left( {{w_t} - {e_t}} \right) \approx F\left( {{w_t}} \right) - \nabla F{\left( {{w_t}} \right)^T}{e_t} + \frac{1}{2}e_t^T{H_t}{e_t}, 
\end{equation}
where ${H_t} = \sum\nolimits_k {{H_{k,t}}}  = \sum\nolimits_k {{\nabla ^2}{F_k}\left( {{w_t}} \right)}$. The Hessian matrix $H_t$ is approximated by its diagonal form $H \approx diag\left( {{h_i}} \right)$,and $e_t$ can be interpreted as a weighted average of all pruned model parameters according to the Eq.~(\ref{eq:global_update_noise}),
\begin{equation}\label{def_error}
  {e_t} = \sum\nolimits_k {{p_k}\left( {1 - {m_{k,t}}} \right) \odot {w_t}} .
\end{equation}
Define $e_{i,t}$ as pruning noise at coordinate $i$ of model $w_t$, and ${e_{i,t}}\! =\! {q_{i,t}}{w_{i,t}}$. Here, ${q_{i,t}}\! =\! \sum\nolimits_k {{p_k}\left( {1 \!-\! {m_{i,k,t}}} \right)} $ and $w_{i,t}$ represents the value associated with model weight at coordinate $i$, then the Eq.~(\ref{loss_error}) can be further simplified as follows:
\begin{equation}\label{final_metric}
  \begin{gathered}
  F\left( {{w_t} - {e_t}} \right) - F\left( {{w_t}} \right) \approx  - \nabla F{\left( {{w_t}} \right)^T}{e_t} + \frac{1}{2}e_t^T{H_t}{e_t} \hfill \\
   =  - \sum\limits_i {{q_{i,t}}{\nabla _i}F\left( {{w_t}} \right){w_{i,t}}}  + \frac{1}{2}\sum\limits_{i,j} {{h_{i,j}}{q_{i,t}}{q_{j,t}}{w_{i,t}}{w_{j,t}}}  \hfill \\
   =  - \sum\limits_i {{q_{i,t}}{\nabla _i}F\left( {{w_t}} \right){w_{i,t}}}  + \frac{1}{2}\sum\limits_i {q_{i,t}^2{h_{i,t}}w_{i,t}^2}.  \hfill \\ 
\end{gathered} 
\end{equation}

\textbf{Remark 2.} This result indicates that the performance of federated model pruning is influenced by three key factors: the model weights, the gradients, and the Hessian matrix. The first term reveals that larger weights and gradients lead to a greater impact on model performance. However, as the number of training rounds increases, the gradient magnitude progressively diminishes, then the curvature information $h_i$ and the weight value of pruned parameters become the dominant factors.

\textbf{Remark 3.} Conventional FL pruning methods typically rely solely on weight-magnitude or gradient-based criteria while neglecting the second-order curvature term $H_t$. When the curvature is substantial (e.g., under non-IID data), using weight norms alone deviates from the optimal pruning strategy, and the resulting residual error is amplified over multiple rounds of aggregation.

Thus, we define an curvature of loss perturbation $s_i$ as a metric for quantifying the importance of each parameter in FL pruning:
\begin{equation}\label{Sig_score}
  {s_{i,t}} = {\nabla _i}F\left( {{w_t}} \right){w_{i,t}} + {h_{i,t}}w_{i,t}^2.
\end{equation}

A smaller $s_{i,t}$ implies that pruning parameter $i$ incurs a smaller increase in the objective value according to the loss analysis in~\eqref{final_metric}, and thus the corresponding parameter is less important. In practice, the server cannot access the exact gradients or Hessian information from all participating clients. To address this limitation, we approximate the gradient and curvature terms using the local model variations, which provides an estimation of parameter importance without additional communication and computation in FL.

\subsection{Heterogeneous Model Pruning and Reconstruction}
In this subsection, we detail how HFLP performs heterogeneous model pruning on the server side, focusing on \emph{mask generation} and \emph{model reconstruction} for convolutional neural networks (CNNs).

\subsubsection{Mask Generation Based on Curvature-Aware Loss Perturbation}
To accommodate heterogeneous client resources, the server assigns client-specific pruning ratios that satisfy communication and computation constraints. Given a pruning ratio, HFLP applies \textbf{hard, structured} pruning by generating a binary mask $m_{k,t}$, where $m_{i,k,t} \!=\! 0$ denotes a pruned parameter and $m_{i,k,t} \!=\! 1$ denotes a retained one. In early rounds, masks are initialized using simple heuristics (e.g., random pruning). As training progresses, mask generation is guided by the curvature-based score in~\eqref{Sig_score}, which can be approximated as $s_i \approx h_i w_{i,t}^2$, when the gradient term becomes small. Parameters with smaller scores are pruned first, yielding masks that minimize loss perturbation while adapting to data heterogeneity and personalized pruning requirements.

\subsubsection{Model Reconstruction for Heterogeneous Sub-Models}
Given the global model $w_t$ and the client-specific mask $m_{k,t}$, the server generates the personalized sub-model for client $k$ via the element-wise masking operation as described in~\eqref{eq:masked_init}. After local training on the pruned sub-model, each client returns an updated heterogeneous sub-model $w_t^{(k,E)}$ to the server. Due to the client-specific masks, these sub-models differ in their effective structures, and thus cannot be directly aggregated in vanilla FedAvg manner.

\begin{figure}[t]
\begin{center}
\includegraphics[width=7.5cm,height=4.8cm]{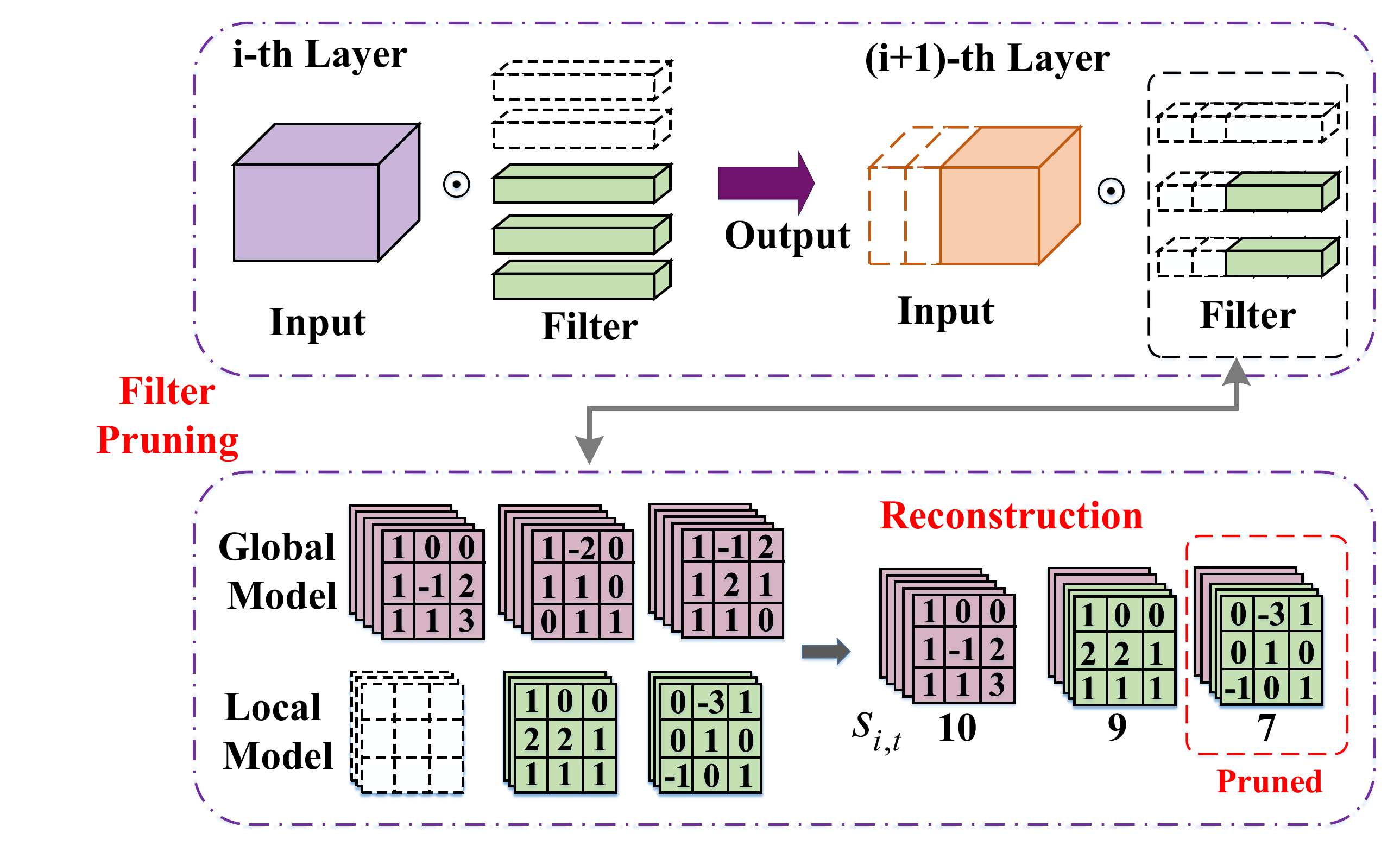}
\caption{Model Pruning and Reconstruction.}
\vspace{-7mm}
\label{reconstruction}
\end{center}
\end{figure}

To address this, the server performs \emph{sub-model reconstruction} prior to aggregation. As illustrated in Fig.~\ref{reconstruction}, each heterogeneous sub-model is expanded to the full model dimension according to the structure of $w_t$: for every pruned filter or neuron, the corresponding entries are filled with the current global parameters at the same indices. In the case of structured pruning in CNNs, pruning a filter in one layer also removes the associated input channel of the corresponding filters in the next layer. During reconstruction, both the removed filters and the associated input channels are restored from the global model. Through reconstructing full-dimensional model $\tilde{w}_t^{(k,E)}$ of client $k$, all processed local models share an identical architecture with the global model.

This reconstruction ensures that the global model combines both its current importance (for parameter where it is active) and its potential in latter round (for parameter where it is pruned), thereby maintaining a fair and consistent aggregation process. After aggregation, the updated global model $w_{t+1}$ is combined with newly generated masks $m_{k,t+1}$ to form the next-round local sub-models.

\section{Experimental Evaluation}
\begin{table*}[htbp]
\centering
\caption{Comparison of CA-HFP with baselines on different datasets (in terms of accuracy \%), $\alpha$ is Dirichlet parameter.}
\begin{tabular}{@{}l | c | c | c | c | c @{\hskip 8pt} c @{\hskip 8pt} c | c @{\hskip 8pt} c @{\hskip 8pt} c | c @{\hskip 8pt} c @{\hskip 8pt} c@{}}
\toprule
\multirow{2}{*}{\makecell[c]{\textbf{Model}}}
& \multirow{2}{*}{\makecell[c]{\textbf{Algorithm}}} 
& \multirow{2}{*}{\makecell[c]{\textbf{Parameters} \\\ Up /Down}} 
& \multirow{2}{*}{\makecell[c]{\textbf{FLOPs}}} 
& \multirow{2}{*}{\makecell[c]{\textbf{Sys.} \\\ \textbf{Heter.}}} 
& \multicolumn{3}{c|}{\makecell[c]{\textbf{FMNIST}}} 
& \multicolumn{3}{c|}{\makecell[c]{\textbf{CIFAR10}}}
& \multicolumn{3}{c}{\makecell[c]{\textbf{CIFAR100}}} \\ 
&&& & & $\alpha$=0.1 & $\alpha$=1.0 & $\alpha$=10.0 & $\alpha$=0.1 & $\alpha$=1.0 & $\alpha$=10.0 & $\alpha$=0.1 & $\alpha$=1.0 & $\alpha$=10.0 \\ 

\midrule
\midrule
\multirow{6}{*}{\makecell[c]{\textbf{VGG}\\\textbf{16}}}
 & FedAvg  &15.1 /15.1 M &40.3 G&$\times$  & 89.02& 93.09& 93.30&73.57 &92.66 &
 92.92& 49.21& 59.79& 61.46\\
 
 & FedProx &15.1 /15.1 M&20.2 G&$\times$ & 85.46& 93.47& 69.48& 74.08& 85.86&
 86.30& 50.01& 53.21& 53.90 \\
 
 & PruneFL &4.30 /4.30 M&20.1 G&$\times$ & 77.08& 92.54 & 92.94& 54.48 & 80.19&
 85.97& 36.80& 39.57& 40.29 \\
 
 & FedMP  &3.95 /3.95 M&10.2 G&$\checkmark$ & 73.34 & 92.47 & 92.79& 54.89 & 76.87&
78.71& 35.97& 41.33& 39.14\\
 
 & DapperFL &3.95 /15.1 M&10.3 G&$\checkmark$ & 59.60& 91.08& 90.74& 51.60 & 79.63&
85.62& 58.32& 46.25& 44.70\\

 & \textbf{CA-HFP}
 &\textbf{3.95 /3.95 M}
 &\textbf{10.2 G}
 &\textbf{$\checkmark$}
 & \cellcolor[rgb]{0.88,0.88,0.95}89.00
 & \cellcolor[rgb]{0.88,0.88,0.95}93.12
 & \cellcolor[rgb]{0.88,0.88,0.95}93.70
 & \cellcolor[rgb]{0.88,0.88,0.95}73.12
 & \cellcolor[rgb]{0.88,0.88,0.95}90.15
 & \cellcolor[rgb]{0.88,0.88,0.95}91.50
 & \cellcolor[rgb]{0.88,0.88,0.95}56.27
 & \cellcolor[rgb]{0.88,0.88,0.95}61.30
 & \cellcolor[rgb]{0.88,0.88,0.95}61.71 \\
 
\midrule
\midrule

\multirow{6}{*}{\makecell[c]{\textbf{Resnet}\\\textbf{32}}}
 & FedAvg  &466 /466 K&9.01 G&$\times$ & 80.80& 92.70& 91.59& 55.71 & 87.43 &
 88.87& 36.37& 39.02& 39.58 \\
 
 & FedProx  &466 /466 K&4.45 G&$\times$ & 86.65& 92.40& 92.51& 58.48 & 79.79 &
 80.54& 35.90& 38.96& 39.92\\
 
 & PruneFL  &261 /261 K&5.45 G&$\times$ & 73.90& 85.13& 81.92& 54.37 & 80.21 &
 81.36& 38.12& 45.51& 41.49 \\
 
 & FedMP   &254 /254 K&5.20 G&$\checkmark$ & 67.49& 80.01& 89.99& 43.55&  67.47&
 69.53& 26.55& 34.46& 32.93  \\
 
 & DapperFL  &254 /466 K&5.23 G&$\checkmark$ & 73.53& 85.11& 88.98&53.18 &80.05 &
 81.21& 37.77& 45.36& 43.35 \\

 & \textbf{CA-HFP}
 &\textbf{254 /254 K}
 &\textbf{5.20 G}
 &$\checkmark$
 & \cellcolor[rgb]{0.88,0.88,0.95}80.90
 & \cellcolor[rgb]{0.88,0.88,0.95}90.13
 & \cellcolor[rgb]{0.88,0.88,0.95}90.59
 & \cellcolor[rgb]{0.88,0.88,0.95}55.98
 & \cellcolor[rgb]{0.88,0.88,0.95}84.61
 & \cellcolor[rgb]{0.88,0.88,0.95}85.03
 & \cellcolor[rgb]{0.88,0.88,0.95}38.41
 & \cellcolor[rgb]{0.88,0.88,0.95}46.25
 & \cellcolor[rgb]{0.88,0.88,0.95}44.70\\
\bottomrule
\end{tabular}
\label{Table:HFP_Perfrmance}
\end{table*}

\subsection{Experiment Setup}
\textbf{Heterogeneity Settings}: In our experiments, we utilize three widely used benchmark datasets in image classification, namely FMNIST, CIFAR-10, and CIFAR-100. To evaluate performance under non-IID data distributions, we adopt a Dirichlet distribution $\text{Dir}(\alpha)$ to partition the training data among participating devices. By adjusting the concentration parameter $\alpha$, we control the degree of data heterogeneity across clients: smaller values of $\alpha$ correspond to more severe non-IID distributions, while larger values yield data partitions closer to the IID setting.

To emulate system heterogeneity under resource constraints, clients are grouped into different \emph{ranks}, each representing a distinct level of computation and communication capability. Specifically, The rank 0 consists of 5 devices with a 25\% pruning ratio and 5 devices without pruning. The rank 1 consists of 5 devices with a 25\% pruning ratio and 5 devices at 50\%. The rank 2 consists of 5 devices with a 50\% pruning ratio and 5 devices at 75\%. The rank 3 represents extremely resource-constrained which consists of 5 devices with a 75\% pruning ratio and 5 devices at 90\%.

\textbf{Model Structure}: To validate the generality and robustness of the proposed method, we conduct extensive experiments using convolution neural networks (CNNs), specifically VGG16 and ResNet56. These models represent two widely adopted yet structurally distinct architectures, enabling a comprehensive evaluation of the effectiveness of our approach. For fair comparison, we uniformly set the target model pruning ratios to ([25\%, 50\%, 75\%]) for all baseline and proposed methods, allowing us to systematically assess the performance under different levels of sparsity.

\textbf{Baselines}: To demonstrate the superiority of the pruning algorithm, the algorithm is compared with some existing SOTA (State-of-the-Art) works:

\begin{itemize}
\item FedAvg \cite{FedAvg}: The classical federated learning framework that performs full model aggregation without model pruning.
\item FedProx \cite{FedProx}: The heterogenous FL framework allows clients to perform different local iterations based on their device capabilities.
\item PruneFL \cite{PruneFL}: A method that incorporates model pruning to reduce communication overhead by transmitting updates with large gradient values.
\item FedMP \cite{FedMP}: An efficient adaptive model pruning framework for FL through $l_1$-norm of filters or neurons. 
\item DapperFL \cite{DapperFL}:A domain adaptive FL employs regularization generated by the pruned model, aiming to learn robust representations across domains.
\end{itemize}

\textbf{Implementation Details}: 
We simulate a cross-device federated learning scenario with ($K=10$) clients, all of which participate in every communication round. The global training process runs for ($T=300$) communication rounds. In each round, every client performs one local epoch with a batch size of 128. We adopt SGD as the local optimizer, with a learning rate of 0.1 and momentum of 0.90. All experiments are implemented in PyTorch and executed on an NVIDIA RTX 3090 GPU.

\subsection{Comparison with the SOTA methods}
\subsubsection{Performance Comparison of Proposed CA-HFP}
Table~\ref{Table:HFP_Perfrmance} reports the test accuracy of CA-HFP and baselines on FMNIST, CIFAR-10, and CIFAR-100 under different Dirichlet partitions. Overall, CA-HFP achieves competitive or superior accuracy across datasets and heterogeneity levels while operating with significantly reduced parameters and FLOPs. On VGG16, CA-HFP matches full-model methods under severe non-IID data ($\alpha=0.1$) and consistently outperforms existing pruning-based baselines as heterogeneity decreases, demonstrating effective mitigation of pruning-induced aggregation bias. On the more challenging CIFAR-100 dataset, CA-HFP maintains stable performance across all $\alpha$ values, highlighting its robustness under coupled system and statistical heterogeneity. Similar trends on ResNet32 further confirm the architecture-agnostic effectiveness of CA-HFP.

\begin{figure}[t]
    \centering
    \begin{subfigure}[b]{0.22\textwidth}
        \includegraphics[width=\textwidth]{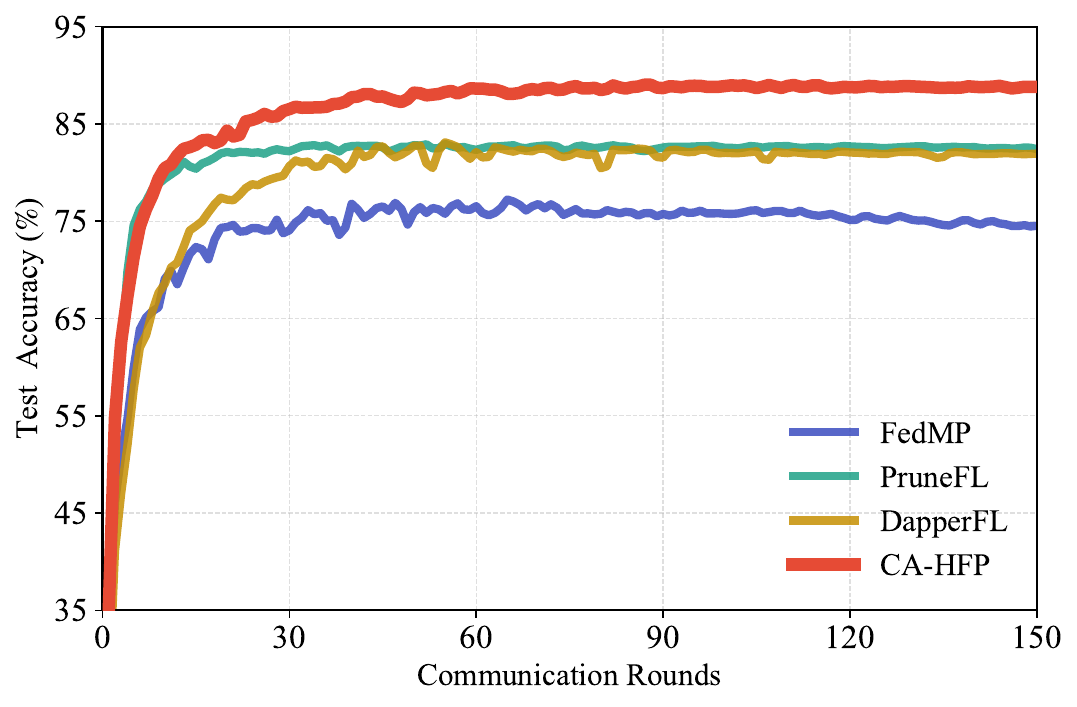}
        \caption{CIFAR10 with $\alpha=$1.0}
        \label{fig:WeightedCofficient}
    \end{subfigure}
    \begin{subfigure}[b]{0.22\textwidth}
        \includegraphics[width=\textwidth]{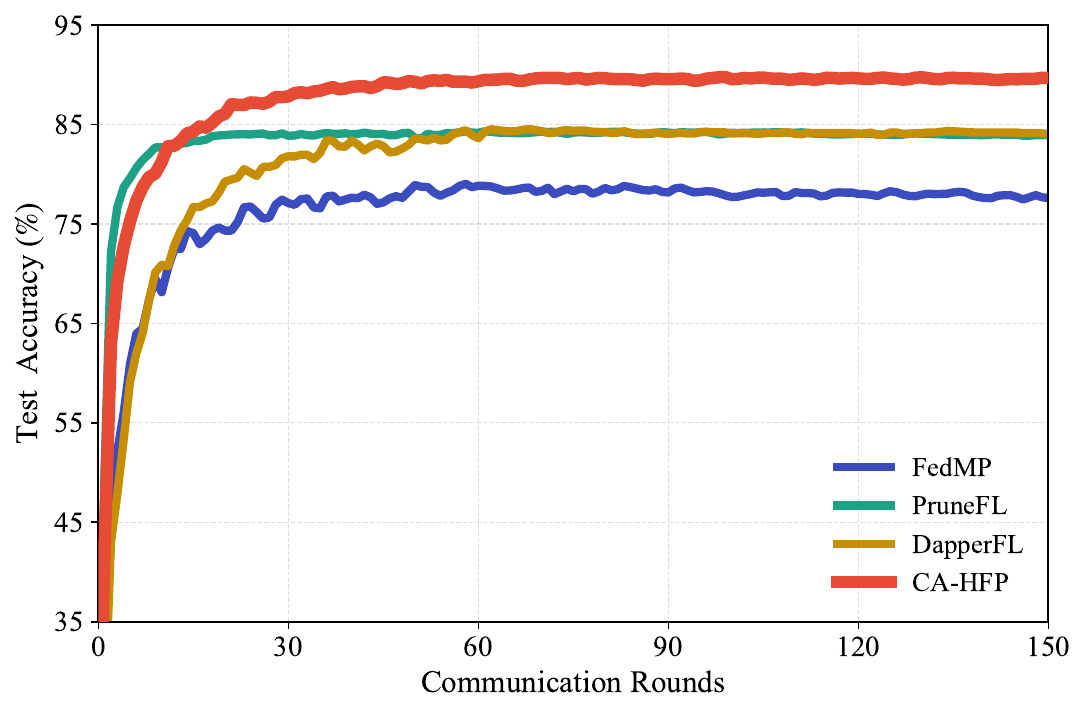}
        \caption{CIFAR10 with $\alpha=$10.0}
        \label{fig:Threshold}
    \end{subfigure}
    \caption{The convergence with different pruning methods.}
    \vspace{-4mm}
    \label{fig:convergence_cifar}
\end{figure}

\begin{figure}[t]
    \centering
    \begin{subfigure}[b]{0.22\textwidth}
        \includegraphics[width=\textwidth]{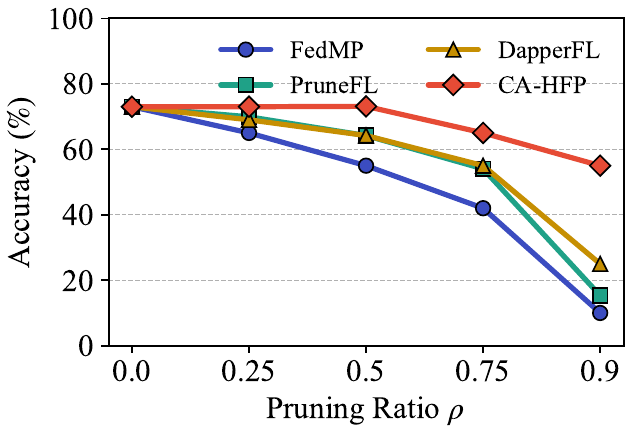}
        \caption{CIFAR10 with $\alpha=$0.1}
        \label{fig:Acc-Ratio-0.1}
    \end{subfigure}
    \begin{subfigure}[b]{0.22\textwidth}
        \includegraphics[width=\textwidth]{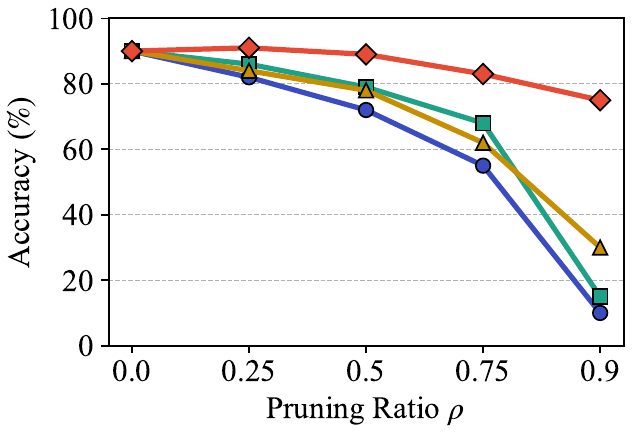}
        \caption{CIFAR10 with $\alpha=$1.0}
        \label{fig:Acc-Ratio-1.0}
    \end{subfigure}
    \caption{The accuracy under different pruning ratio.}
    \vspace{-5mm}
    \label{fig:acc_compare_pruning_ratio}
\end{figure}

\begin{figure}[t]
    \centering
        \begin{subfigure}[b]{0.22\textwidth}
        \includegraphics[width=\textwidth]{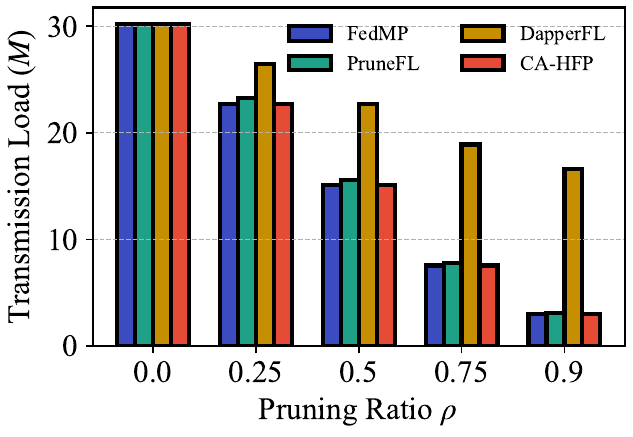}
        \caption{Communication Cost}
        \label{fig:Communication_Cost}
    \end{subfigure}
    \begin{subfigure}[b]{0.22\textwidth}
        \includegraphics[width=\textwidth]{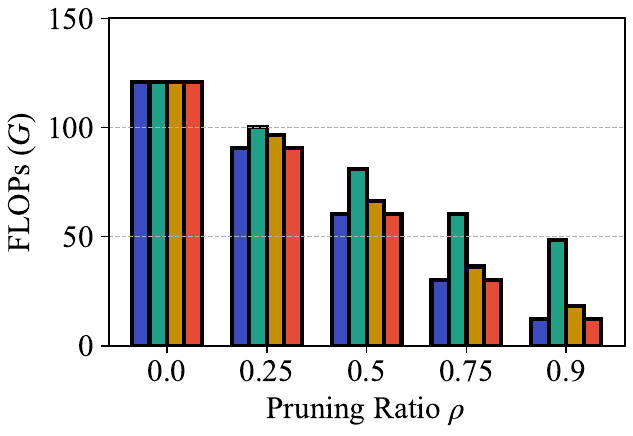}
        \caption{Computing Cost}
        \label{fig:Computing_Cost}
    \end{subfigure}
    \caption{\small{The cost under different pruning ratio. The communication cost includes the uploading and the downloading and the calculation of FLOPS is according to \cite{PruneFL}.}}
    \vspace{-3mm}
    \label{fig:cost_compare_pruning_ratio}
\end{figure}

\begin{table}[t]
\centering
\caption{Comparison of HFP w and w/o reconstruction.}
\begin{tabular}{@{}l| c | c @{\hskip 8pt} c @{\hskip 8pt} c | c @{\hskip 8pt} c @{\hskip 8pt} c @{}}
\toprule

\multirow{2}{*}{\makecell[c]{\textbf{Metric}}}
& \multirow{2}{*}{\makecell[c]{\textbf{Recon.}}}
& \multicolumn{3}{c}{\makecell[c]{\textbf{CIFAR10}}} 
& \multicolumn{3}{c}{\makecell[c]{\textbf{CIFAR100}}} \\ 
& & $\alpha$=0.1 & $\alpha$=1.0 & $\alpha$=10.0 & $\alpha$=0.1 & $\alpha$=1.0 & $\alpha$=10.0  \\ 
\midrule
\midrule

\multirow{2}{*}{\makecell[c]{$l_1$} norm}
& w   & 54.83& 86.81& 88.69 & 45.96 & 46.23 & 49.11\\
& w/o  & 54.74& 84.34& 84.55 & 44.14 & 45.94 & 46.91 \\
\midrule
\multirow{2}{*}{\makecell[c]{$\Delta w$}}
& w   & 54.89& 80.16& 85.79 & 46.21 & 49.13 & 50.31 \\
& w/o  & \emph{False}& --& -- & -- & -- & -- \\
\midrule
\multirow{2}{*}{\makecell[c]{Curv.}}
& w   & \textbf{73.12}& \textbf{90.15}& \textbf{91.50} & \textbf{56.27} & \textbf{61.30} & \textbf{61.71} \\
& w/o  & 62.84& 88.96& 90.37 & 33.44 & 44.24 & 45.93 \\
 
\bottomrule
\end{tabular}
\vspace{-3mm}
\label{Table:Aggregations_Study}
\end{table}
\begin{figure}[t]
    \centering
        \begin{subfigure}[b]{0.22\textwidth}
        \includegraphics[width=\textwidth]{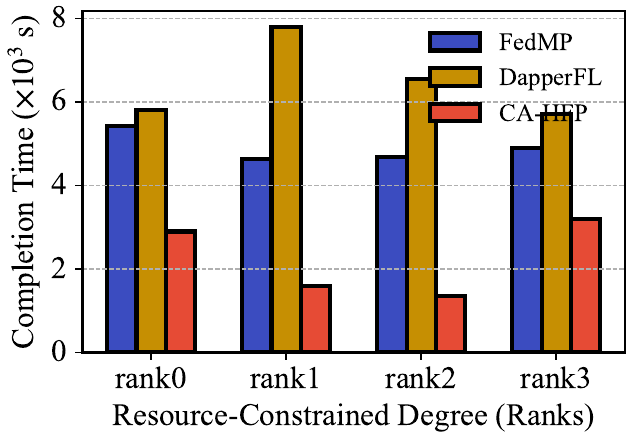}
        \caption{Completion Time.}
        \label{fig:Heterogeneity_ACC}
    \end{subfigure}
    \begin{subfigure}[b]{0.22\textwidth}
        \includegraphics[width=\textwidth]{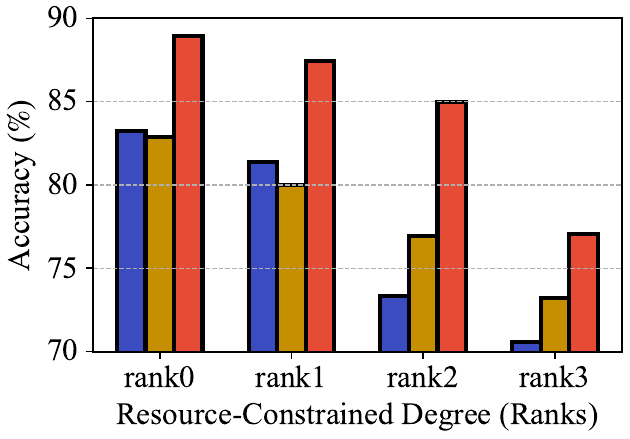}
        \caption{Accuracy}
        \label{fig:Heterogeneity_Time}
    \end{subfigure}
    \caption{\small{The performance under different heterogeneous system}}
    \vspace{-6mm}
    \label{fig:performance_heterogeneity}
\end{figure}

\subsubsection{Convergence of CA-HFP}
Fig.~\ref{fig:convergence_cifar} compares the convergence behavior of different pruning-based FL methods on CIFAR-10 under non-IID data distributions. Under both moderate ($\alpha=1.0$) and mild ($\alpha=10.0$) heterogeneity, CA-HFP converges noticeably faster than FedMP, PruneFL, and DapperFL, while consistently achieving the highest final accuracy. This improvement indicates that CA-HFP effectively alleviates aggregation bias induced by heterogeneous pruning and non-IID data. Overall, the results demonstrate that CA-HFP offers a superior accuracy–communication trade-off and robust convergence across different heterogeneity levels.

\subsubsection{Impact of Pruning Degree on CA-HFP}
Fig.~\ref{fig:acc_compare_pruning_ratio} illustrates the accuracy performance under different pruning ratios $\alpha$ on CIFAR-10. Under severe non-IID data ($\alpha=0.1$), CA-HFP consistently outperforms all baselines across all pruning levels, with the performance gap widening as pruning becomes more aggressive, demonstrating strong robustness to compounded statistical heterogeneity and model sparsification. Under milder heterogeneity ($\alpha=1.0$), CA-HFP still achieves the highest accuracy, particularly at high pruning ratios (e.g., $\rho=0.75$ and $\rho=0.9$). These results indicate that CA-HFP maintains stable performance even under substantial model compression and heterogeneous data distributions.

\subsubsection{Communication and Computing Cost} Fig.~\ref{fig:cost_compare_pruning_ratio}(a) shows that increasing $\rho$ significantly reduces the transmission load for CA-HFP, yielding markedly lower communication cost at different pruning levels compared with the baselines, while Fig.~\ref{fig:cost_compare_pruning_ratio}(b) further confirms that CA-HFP achieves corresponding reductions in computation cost (TFLOPs) as $\rho$ increases. Overall, Fig.~\ref{fig:cost_compare_pruning_ratio} validates that CA-HFP provides a favorable accuracy--efficiency trade-off: it preserves superior accuracy under non-IID data while substantially reducing both communication overhead and local computation through structured and hard pruning.

\subsubsection{Effect of Resource Constraints}
Fig.~\ref{fig:performance_heterogeneity} evaluates the performance of different pruning-based FL methods under heterogeneous system constraints characterized by client resource ranks. As shown in Fig.~6(a), CA-HFP consistently achieves the shortest completion time across all resource ranks, demonstrating its efficiency advantage in resource-constrained systems. Meanwhile, Fig.~\ref{fig:performance_heterogeneity}(b) shows that CA-HFP maintains the highest test accuracy under all system settings, while baseline methods suffer noticeable degradation as resource constraints increases. These results indicate that CA-HFP effectively balances efficiency and accuracy, enabling robust federated training under heterogeneous system.

\subsubsection{Ablation Study for Reconstruction}
Table~\ref{Table:Aggregations_Study} reports the ablation results of CA-HFP with and without the server-side reconstruction step. Reconstruction brings the most significant gains under severe non-IID settings: on CIFAR-10 with $alpha=0.1$, it improves accuracy from 62.84\% to 73.12\%, highlighting its critical role in mitigating aggregation bias caused by heterogeneous pruning. Similar improvements are observed on CIFAR-100, where reconstruction substantially stabilizes training and improves robustness across all heterogeneity levels. These results confirm that reconstruction effectively aligns heterogeneous sparse models, enabling reliable aggregation under misaligned client masks.

\section{Conclusion}
\label{sec:conclusion}

This paper presents CA-HFP, a practical framework that adapts federated learning to heterogeneous environments by enabling device-specific structured pruning while preserving global aggregation through a lightweight server-side reconstruction. By quantifying the effects of data heterogeneity and pruning-induced perturbations in the convergence analysis, CA-HFP derives a loss curvature–informed importance score for personalized compression across heterogeneous clients. Experiments on diverse datasets and models show that the proposed framework achieves higher accuracy and significant reductions in per-client computation and communication costs, while robustly adapting to heterogeneous environments.

\bibliographystyle{IEEEbib}
\bibliography{icme2026references}

\end{document}